\RequirePackage{fix-cm}
\documentclass[twocolumn]{arxiv}
\smartqed  
\usepackage{graphicx}

\pdfoutput=1

\usepackage{graphicx,amsmath,url}      % include this line if your document contains figures
\usepackage[round]{natbib}             % required for bibliography
\usepackage[english]{babel}
\usepackage{amsmath}

\usepackage[dvipsnames]{xcolor}

\usepackage{makeidx}         % allows index generation
\usepackage[bottom]{footmisc}% places footnotes at page bottom
\usepackage{graphics}
\usepackage{times} % assumes new font selection scheme installed
\usepackage{amsmath,lipsum} % assumes amsmath package installed
\usepackage{amssymb}  % assumes amsmath package installed
\usepackage{mathrsfs}
\usepackage{color}
\usepackage{comment}
\usepackage{float}
\usepackage{enumerate}
\usepackage{marvosym }
\usepackage{array}
\usepackage{color}

\usepackage{etoolbox} % quebra de linha url
\gappto{\UrlBreaks}{\UrlOrds}% quebra de linha url

\usepackage{hyperref}

\usepackage{amssymb}
\usepackage{mathtools}

\usepackage{soul}
\usepackage[ruled,vlined]{algorithm2e}
\usepackage[noend]{algpseudocode} % 
\definecolor{adriano}{RGB}{255, 100, 0}

\definecolor{victor}{RGB}{0, 40, 200}

\definecolor{hector}{RGB}{10, 150, 200}

\definecolor{luciano}{RGB}{50, 150, 50}

\begin{document}

%Space above and below the equations
\setlength{\abovedisplayskip}{5pt}
\setlength{\belowdisplayskip}{5pt}

\title{Autonomous Navigation System for a Delivery Drone
}

\author{
Victor R. F. Miranda$^{1}$
\and
Adriano M. C. Rezende$^{1}$
\and
Thiago L. Rocha$^{1}$
\and
H\'ector Azp\'urua$^{2,3}$\\
\and
Luciano C. A. Pimenta$^{1}$
\and
Gustavo M. Freitas$^{1}$
}

\institute{$^{1}$ 
              Escola de Engenharia, Universidade Federal de Minas Gerais. \email\texttt{\{victormrfm, adrianomcr, thiagolages, lucpim, gustavomfreitas\}@ufmg.br}\\
           $^{2}$ 
              Dept. de Ciência da Computação, Universidade Federal de Minas Gerais. \email{\texttt{hector.azpurua@dcc.ufmg.br}}\\
            $^{3}$
              Instituto Tecnológico Vale (ITV) - Mining. \email{\texttt{hector.azpurua@itv.org}}
              }

\date{}

\color{black}

\maketitle

\begin{abstract}
The use of delivery services is an increasing trend worldwide, further enhanced by the COVID pandemic. 
In this context, drone delivery systems are of great interest as they may allow for faster and cheaper deliveries.
This paper presents a navigation system that makes feasible the delivery of parcels with autonomous drones. %to allow the feasible delivery of parcels by autonomous drones.
The system generates a path between a start and a final point and controls the drone to follow this path based on its localization obtained through GPS, 9DoF IMU, and barometer. In the landing phase, information of poses estimated by a marker (ArUco) detection technique using a camera, ultra-wideband (UWB) devices, and the drone's software estimation are merged by utilizing an Extended Kalman Filter algorithm to improve the landing precision. A vector field-based method controls the drone to follow the desired path smoothly, reducing vibrations or harsh movements that could harm the transported parcel.
Real experiments validate the delivery strategy and allow to evaluate the performance of the adopted techniques. 
Preliminary results state the viability of our proposal for autonomous drone delivery.

\end{abstract}

\section{Introduction}
\label{sec:intro}

Autonomous robots have been studied for a long time, and the rising demand for automated solutions for real-world problems has accelerated research in the area. Such problems involve housekeeping tasks (vacuum and lawnmower robots), military missions (rescue, patrol, attacks), security applications (surveillance, exploration), industrial operations (production and logistics), among others.

In the past few years, the online shopping market has grown significantly, and ever since, retail and mail companies seek to make autonomous drone delivery a reality. The global online food delivery services reached \$136.4bn in 2020, a 27\% increase from the same period in 2019~\citep{AJOT}. This increase should continue in the following years with an expectation of \$182.3bn in 2024.
As a result, delivery services face a high demand for orders and sometimes cannot maintain a short delivery time. 
Thus, new and innovative means of transportation  %forms
must be developed to increase efficiency and cope with the increasing demand. 

%must continue year by year 

%In this scenario, 
Autonomous delivery presents itself as a very convenient alternative, particularly in social isolation periods, such as those experienced by many countries in 2020 and 2021 due to the COVID-19 pandemic. In situations like this, physical contact between people should be reduced as much as possible, especially with people from risk groups. Therefore, the transportation of goods by autonomous quadrotors could prevent any physical contact with the customer, thus maintaining the World Health Organization (WHO) recommendations and preventing the proliferation of the virus.

Some of the world's biggest companies, such as Amazon, UPS, and Alphabet, are making advances toward drone delivery services. According to~\cite{schneider_2020}, FlightForward, the UPS branch responsible for drone flights, has already achieved air carrier certification, which allows it to deliver small packages with drones. Wing, a division of Alphabet, launched the United States' first small commercial delivery service in Christiansburg, Virginia. These achievements happened in late 2019 and show that the market for drone delivery is undoubtedly gaining ground worldwide. In late 2020, Brazil's National Aviation Agency (ANAC) granted the first authorization to a private company, SpeedBird, to perform drone cargo tests in Brazillian urban areas \citep{anac2021drone}. 

In this context, %sense,
this paper presents a methodology for enabling autonomous drone deliveries. After generating a path between two points of interest, the drone's location is obtained through its GPS (Global Positioning System), 9DoF IMU (Inertial Measurement Unit), and a barometer. A vector field control algorithm then guides the quadrotor into the desired path. However, in drone delivery, reproducible safe landings in urban areas are a critical challenge. In that respect, we propose an Extended Kalman Filter (EKF) algorithm that fuses planar visual marker and Ultra-Wideband (UWB) localization strategies with the drone's software pose %fusion 
estimation to improve landing accuracy. 
The visual localization uses the ArUco markers
% \footnote{Augmented Reality Marker - \href{https://www.uco.es/investiga/grupos/ava/node/26}{https://www.uco.es/investiga/grupos/ava/node/26}}
, and the UWB localization is estimated via multilateration with multiple UWB anchors over the landing area. The proposed method explores the techniques described in \cite{xquadicar}, initially developed for high-performance autonomous drone racing, to create a practical and robust real-world system for drone delivery services. 
%
% Real and simulated experiments validate the feasibility of the proposed strategies.
Real experiments validate the feasibility of the proposed strategies. 

%%%% -------------
%%%% START ORIGINAL
%%%% -------------
% \textcolor{green}{This paper presents a practical application of some techniques previously deployed by the XQuad-UFMG team for an autonomous drone race competition~\citep{xquadicar}, \textcolor{red}{with adaptations and extensions considering} delivery services.}

% \victor{
% In this context, this paper presents a methodology for enabling deliveries by an autonomous drones}. After generating a path between two points of interest, the drone's location is obtained through its GPS (Global Positioning System), IMU (Inertial Measurement Unit) and barometer. Once its location is obtained, the control of the quadrotor is performed using a vector fields technique, which makes the drone follow the desired path. 
% %
% \victor{
% To improve landing accuracy, we have developed an Extended Kalman Filter (EKF) algorithm that fuse localization information from a multilateration algorithm using Ultra-wideband (UWB) devices, a marker detection and Perspective-n-Point (PNP) technique using a camera pointing to an ArUco marker\footnote{Augmented Reality Marker - \url{https://www.uco.es/investiga/grupos/ava/node/26}} and the drone's original fusion estimation (GPS,IMU and barometer).
% }
% % To improve landing accuracy, Ultra-wideband (UWB) devices or a camera pointing to an ArUco marker\footnote{Augmented Reality Marker - \url{https://www.uco.es/investiga/grupos/ava/node/26}} were used. 
% Simulations and real experiments show that the strategies used are feasible to carry out autonomous package deliveries.
%%%% -------------
%%%% END ORIGINAL
%%%% -------------

The remainder of the paper is structured as follows. First, the related works are presented in Section~\ref{sec:related}. Section~\ref{sec:problem} describes the problem of delivery with autonomous drones, and Section~\ref{sec:methodology} presents the methodologies used to accomplish this task. The results obtained are discussed in Section~\ref{sec:results}. Finally, conclusions are presented in Section~\ref{sec:conclusions}, together with future research perspectives.

\section{Related Works}
\label{sec:related}

Given the increasing demand involving autonomous air transport of cargo over short distances, several studies address different strategies for load transportation and other common problems in this type of task. Drone delivery is an emerging field, gaining attention in the academy and industry given the numerous challenges to overcome to perform successful missions in urban environments, such as guidance, trajectory planning, control, localization, obstacle avoidance, and safe landing, especially when global localization is not available or is unreliable~\citep{yoo2018drone}. Parcel delivery using drones is also gaining attention, given the environmental benefit of aerial platforms against standard truck delivery~\citep{koiwanit2018analysis}. 
% \textcolor{red}{In this sense, \cite{villa2019survey} presented a survey revision addressing meaningful research works of load transportation using multi-rotor Unmanned Aerial Vehicles (UAVs).}

%%%%%%%%%%%%%%%%%%%%%%

Regarding the control methods for load transportation applications, \cite{raffo2016} propose a robust nonlinear control technique for load transportation using quadrotors. Despite proving asymptotic stability, they consider a cable-suspended transport system susceptible to external disturbances due to the wind %winds
and drone maneuvers. Similarly, \cite{zuniga2018load} present cooperative cable-suspended load transportation, using multiples drones with consensus strategies. This approach reduces in 60\% the cable oscillations. Unlike these approaches, we use a transportation system with the load attached to the drone's body, reducing disturbances due to the cargo movements, and a control method that minimizes aggressive maneuvers when following the path.

Several other control strategies are present in the literature for autonomous drone operations. However, most of them focus on ensuring the drone's stability at the lowest level, acting on the motors to follow a trajectory \citep{almakhles2019robust}. The present work considers that the drone already has the lowest level controller properly tuned to guarantee the desired angular velocities. Therefore, we adopt the approaches presented in \cite{9196605} and \cite{goncalves2010} to define our high-level control strategy, also commonly called guidance. \cite{9196605} propose a nonlinear path control method based on artificial vector fields that consider the robot's dynamics. \cite{goncalves2010} present a theory to generate the vector field in $\mathbb{R}^n$, which can be used considering the three-dimensional case.

%%%%%%%%%%%%%%%%%%%%%%

Recent works dealing with drone delivery have focused on route optimization~\citep{chiang2019impact}, optimal charging station location~\citep{hong2018range}, and the mixture of traditional aerial routes with drone-carrying truck routes \citep{chang2018optimal,boysen2018drone}. However, a holistic analysis of the requirements of a suitable delivery platform is often overlooked. Motivated by this, we propose a navigation system for complete autonomous delivery tasks using drones, focusing on the practical aspects of safe drone landing.

Traditional protocols and drone landing methods rely on expensive equipment such as DGPS or RTK GPS or do not satisfy the precision and robustness needed for drone landings in urban areas. Visual localization methods could aid in locating the landing area accurately, even in partially cluttered scenarios, using equipment already deployed in the platform, for instance, RGB cameras. Planar markers such as the ArUco \citep{garrido2014automatic}, can generate robust pose estimation from heights up to 30 m and are a feasible solution for drone landing~\citep{wubben2019accurate,marut2019aruco}. Many landing solutions using planar markers change the flight behavior to use only ArUco localization when in range instead of performing fusion sensing. Despite the low-cost, low-power consumption of visual localization using planar markers, they are more prone to environmental interference such as low light conditions, snow, dust, rain, and fog. Therefore, visual planar markers require constant maintenance to keep them fully functional. A different visual approach for robust landing in vision-compromised environments uses active Infrared (IR) beacons located at the landing platform and a special IR camera for detecting them from afar~\citep{nowak2017development}. These methods can work without external illumination but could be imprecise at direct sunlight or other IR emission sources.

Other types of localization systems that are less prone to environmental interference are wireless-based localization systems. Robust wireless localization can be achieved with Ultra-Wideband (UWB) systems using the Time-of-Flight (ToF) principle to estimate distances with centimeter precision. Drones could exploit these localization systems for indoor localization in cluttered environments such as in \cite{tiemann2017scalable} and \cite{tiemann2018enhanced}. Key benefits of these types of wireless localization technologies are that they could work even in visually degraded situations, are easily scalable to multiple platforms~\citep{nguyen2016ultra}, and are particularly robust to walls and reflections, increasing the possible range of real-world situations where they can be applied. Recent works have fused UWB and vision localization for drone landing based on recursive least square optimization~\citep{nguyen2019integrated}.

Our work differs from the previously mentioned ones since we propose a complete platform for drone delivery and compare popular localization methods for UAV platforms such as GPS, visual, and UWB localization in the landing phase. We also propose a sensory fusion of multiple external localization techniques given the sensing capabilities already available on the UAV, including GPS, 9DoF IMU, and barometer. An Extended Kalman Filter improves landing accuracy considering fixed location platforms in urban areas.

%\vspace{-0.1cm}
\section{Autonomous Drone Delivery}
\label{sec:problem}

%\vspace{-0.2cm}

This section describes the autonomous delivery problem with drones and specifies the hardware and software used for development.

%  software and the simulation environment used for development.

\begin{figure}[t]
      \hspace{-0.5cm}
      \centering
      \includegraphics[clip,trim={0.0in 0.0in 0.0in 0.0in},width=0.485\textwidth]{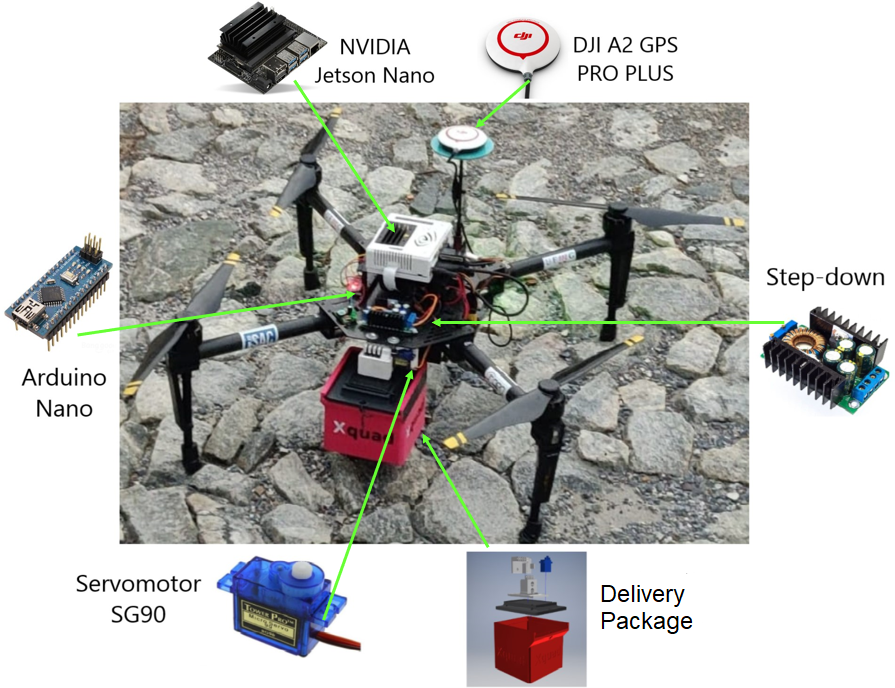}
      \vspace{-0.0cm}
      \caption{Drone used for the autonomous delivery and its components.}
      \label{fig:drone_completo}
      \vspace{-0.0cm}
\end{figure}

\subsection{Problem Description}

The problem addressed involves transporting small parcels between two points of interest using a drone in autonomous mode without receiving commands from a human pilot.
We have considered obstacle-free environments and assumed distances compatible with the drone endurance (maximum time of flight).
Besides, since it is essential for delivery drones to carry fragile objects without abrupt movements, we have adopted a grasped transportation mechanism that reduces the risks of vibrations or unexpected package drops.

At first, residential regions will be the primary landing locality for delivery tasks; therefore, the drone must have the ability to land with accuracy in restricted and narrow areas to prevent unexpected accidents or injuries. The tests were performed considering a 1x1\textrm{m} landing platform.

% In addition, it is important that the drone can carry relatively fragile objects without abrupt movements, as seen in~\cite{raffo2016} which considers a suspended load. This approach differs from the one used here, as the package is attached to the body of the drone and therefore it cannot swing freely.

%\vspace{-0.2cm}
%\subsection{Resources}

\subsection{Hardware} 

The DJI Matrice 100\footnote{DJI Matrice 100 - \href{https://www.dji.com/br/matrice100}{https://www.dji.com/br/matrice100}}, illustrated in Figure~\ref{fig:drone_completo}, allows developers to implement their codes to control the drone. All sensors that come with the drone by default are used, including GPS, 9DoF IMU, and a barometer. These are responsible for helping to estimate the drone's position in global coordinates (latitude and longitude), orientation, and height. 
% \victor{
The drone's flight board is responsible for transmitting information from the sensors to another device, in addition to receiving control commands and sending them to the brushless motors.
% }
% Along with these sensors, the drone's own flight board interfaces the signals generated by the control strategy and the brushless motors commands. 
In addition to the flight controller board already available on the drone, a Jetson Nano\footnote{Jetson Nano - \href{https://www.nvidia.com/en-us/autonomous-machines/embedded-systems/jetson-nano/}{https://www.nvidia.com/en-us/autonomous-machines/embedded-systems/jetson-nano/}} is responsible for data processing, path planning, and the quadrotor high-level control.

\begin{figure}[htbp]
    \centering
    % \subcaptionbox{}{
        \centering
        \includegraphics[width=0.49\linewidth]{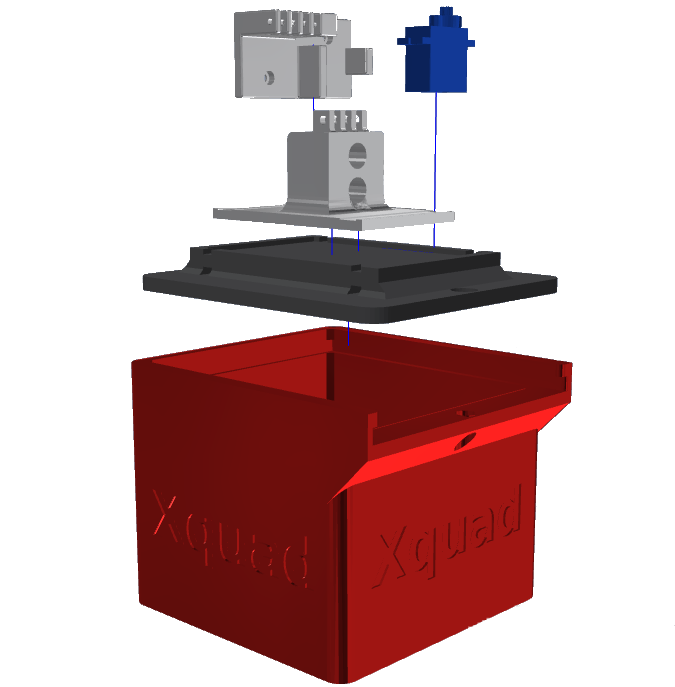}
    % }%
    % \subcaptionbox{}{
        \centering
        \includegraphics[width=0.45\linewidth]{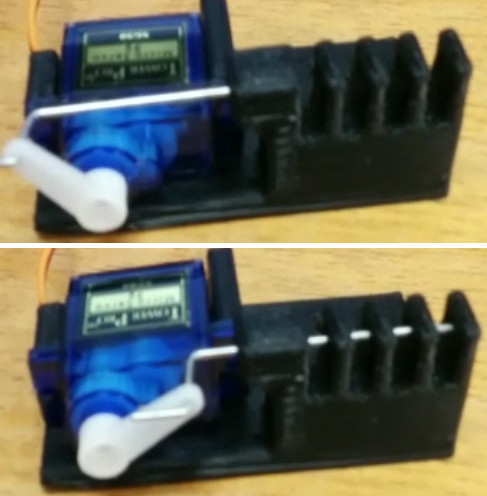}
    % }
    \caption{Coupling mechanism using a servo motor with a package used for cargo delivery: (a) exploded delivery box, and (b) the servo coupling mechanism.}
    \label{fig:exploded_view}
\end{figure}

\begin{figure}[htbp]
      \vspace{-0.0cm}
      \centering
      %\hspace{-0.4cm}
      \includegraphics[width=0.42\textwidth]{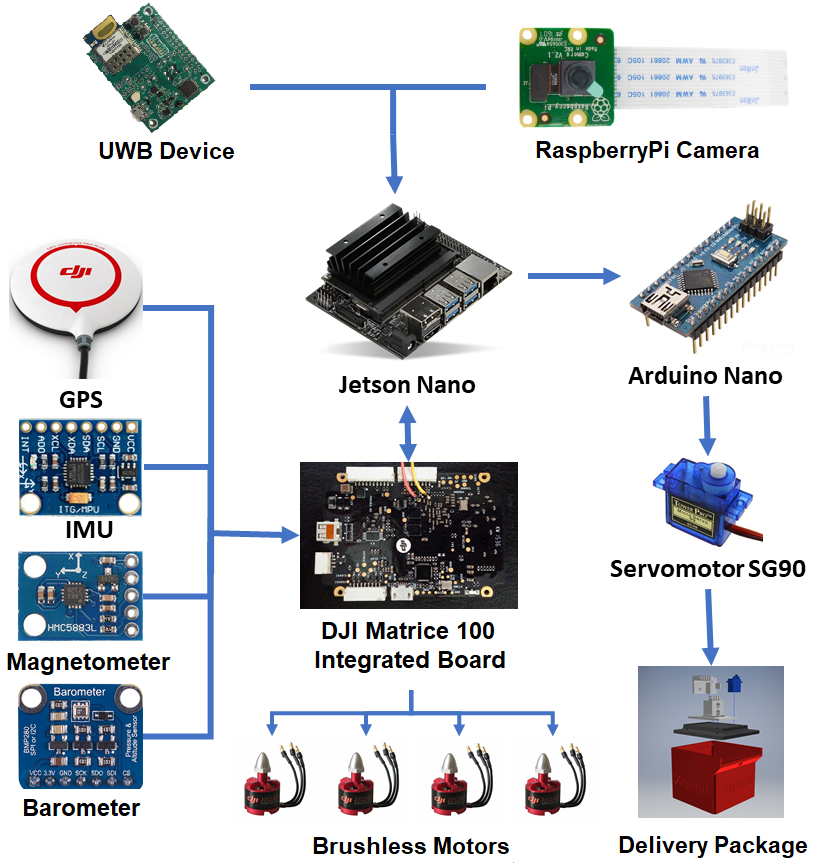}
      \vspace{-0.0cm}
      \caption{Connection between the drone embedded equipment.}
      %High-level representation of the connection between equipment used in the drone.
      \label{fig:esquema_ligacoes}
      \vspace{-0.0cm}
\end{figure}

We also developed a 3D model of the box used for delivery, as shown in Figure~\ref{fig:exploded_view}a. Its coupling mechanism uses a servo motor, as illustrated in Figure~\ref{fig:exploded_view}b. An Arduino Nano is connected to the Jetson board to control this servo motor, placing it in the position of coupling or decoupling the box on the drone. The schematic of Figure~\ref{fig:esquema_ligacoes} illustrates the connections between the equipment and drone used during experiments.

Considering that the landing site used in the experiments has 1x1 \textrm{m} and that the horizontal accuracy in the drone's location using GPS is approximately $2\mathrm{m}$, it is necessary to use additional sensors to assist the drone in making a safer and more accurate landing. %\textcolor{red}{Due to the impossibility of carrying out new tests with the autonomous drone, the application of the following sensors was verified only in a simulated environment}: 
For that, we verified the application of the following sensors
in experiments using the real drone: (i) a RaspberryPi Camera v2.0 pointing downwards, together with ArUco markers combination on the landing platform, and (ii) Ultra-wideband (UWB) devices anchored at the landing site, and a device of the same type attached to the drone. In both cases, it is possible to obtain additional information on the drone's position with respect to the platform during landing and improve the localization by fusing all these data.
%\textcolor{red}{The experimental validation of the use of these sensors will be done as soon as the drone is available for further tests.}

%Duas opções de sensores são então utilizadas em simulação: (i) uma \textit{RaspberryPi Camera v2.0} apontada para baixo juntamente a um marcador ArUco na plataforma ou (ii) dispositivos \textit{Ultra-wideband (UWB)} ancorados no local de pouso, e um dispositivo do mesmo tipo acoplado ao drone. De ambas as formas, é possível obter informações adicionais da posição do drone em relação à plataforma durante o pouso.
%\textcolor{red}{O uso destes dois sensores foi validado de forma preliminar em simulações, e portanto não foi acoplado fisicamente ao drone, como os componentes da Figura \ref{fig:esquema_ligacoes}. }
%

% \begin{figure}[htbp]
%       %\vspace{-0.0cm}
%       \centering
%       \includegraphics[width=0.27\textwidth]{images/exploded_view_small.png}
%       \vspace{-0.0cm}
%       \caption{Coupling mechanism using a servo motor with a package used for cargo delivery.}
%       \label{fig:exploded_view}
%       \vspace{-0.0cm}
% \end{figure}

% \begin{figure}[htbp]
%       \centering
%       \includegraphics[width=.48\textwidth]{images/servo_aberto_e_fechado.png}
%     \caption{Coupling and uncoupling device present in the drone.}
%     \label{fig:servos}
% \end{figure}

\subsection{Software}

The Operating System used is Ubuntu 18.04 together with ROS 1 (Robot Operating System). The ROS package provided by DJI called Onboard-SDK-ROS\footnote{Onboard-SDK-ROS - \href{https://github.com/dji-sdk/Onboard-SDK-ROS}{https://github.com/dji-sdk/Onboard-SDK-ROS}} establishes communication between the drone and the Jetson Nano. The package allows sending commands to the embedded control system and provides data from the sensors present on the drone, such as GPS, 9DoF IMU, and barometer, in addition to estimating the drone's position and orientation in an online fashion. For the identification and estimation of the ArUco's pose, specific algorithms from the OpenCV\footnote{OpenCV - \href{https://opencv.org/}{https://opencv.org/}} are used. In the case of UWB devices, an algorithm uses the difference between the time of arrival (TDoA) of the signal for each device to compute the drone's position with respect to the landing site.

\section{Methodology}
\label{sec:methodology}

\vspace{-0.0cm}

\color{black}

In order to satisfy the problem requirements, such as reducing the delivery time while avoiding abrupt movements and with a safe landing, we propose a solution divided into three distinct tasks: (i) path planning, (ii) localization, and (iii) control.
% In order to satisfy the problem requirements, such as reducing the delivery time while avoiding abrupt movements and collisions, we propose a solution divided into three distinct tasks: (i) path planning, (ii) localization, and (iii) control.

\vspace{-0.0cm}

\subsection{Path Planning} \label{sec:path_planning}
The proposed path planning strategy simplifies autono-mous drone delivery. First, the method considers the altitude, latitude, and longitude data to define the drone's geographical location.
For test purposes, the covered distances are short, such that a flat Earth model can be considered. Thus, we transform the angles of latitude and longitude into measurements of distance. The drone's position is then initially represented with respect to the Earth's reference frame $\mathcal{F}_E$.

Consider that the drone's starting point is $\mathbf{p}_s\in\mathbb{R}^3$ and that the load delivery point is $\mathbf{p}_f\in\mathbb{R}^3$.
Without loss of generality, it is possible to assume an inertial coordinate system $\mathcal{F}_I$ that respects two conditions:
% {\color{adriano}
\begin{enumerate}
    \item The path's end point is the origin, i.e.\, $\mathbf{p}_f = \mathbf{0}$;
    \item The $\mathbf{x}$ axis of the coordinate system $\mathcal{F}_I$ is in the the horizontal plane, pointing in the direction of the final point, i.e.\, $\hat{x} \parallel \Pi_{xy}(\mathbf{p}_f-\mathbf{p}_s)$, where $\Pi_{xy}(\cdot)$ represents the projection in the $\mathbf{xy}$ plane.
\end{enumerate}
The inertial reference frame $\mathcal{F}_I $ is easily obtained through two operations: a translation with respect to $\mathcal{F}_E$, in order to satisfy condition 1; and a simple rotation in the $\mathbf{z}$ axis to satisfy condition 2.

\begin{figure}[t]
      \vspace{-0.0cm}
      \centering
      \includegraphics[clip,trim={1.7in 3.3in 1.7in 3.5in},width=0.4\textwidth]{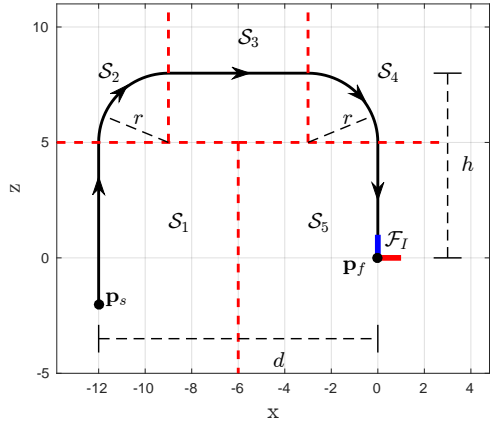}
      \vspace{-0.0cm}
      \caption{Sectors $\mathcal{S}_1$, $\mathcal{S}_2$, $\mathcal{S}_3$, $\mathcal{S}_4$ e $\mathcal{S}_5$, defined in equation~\eqref{eq:sectors}.}
      \label{fig:sectors}
      \vspace{-0.0cm}
\end{figure}

The proposed path planning method computes a smooth reference path connecting the two points $\mathbf{p}_s$ and $\mathbf{p}_f$.
The strategy consists of creating $5$ path sections: (i) vertical ascending line; (ii) arc of a circle; (iii) horizontal line towards the platform; (iv) arc of a circle; (v) vertical descending line.
In order to allow a smooth transition between sections, space is divided into $5$ sectors $\mathcal{S}_i, \ i = 1,2,3,4,5 $. Each path section is associated with one sector.
%\victor{Therefore, the space is divided into $5$ sectors $\mathcal{S}_i, \ i = 1,2,3,4,5 $.}
% Instead of defining a state machine as in the previous case, the space was divided into $5$ sectors $\mathcal{S}_i, \ i = 1,2,3,4,5 $.
%
%\textcolor{red}{Each of them has an associated section that will serve as reference to the drone, depending on its location.}\adriano{[não entendi]} 
The definition of the sectors is presented below:
%
% {\color{adriano}
\begin{eqnarray}
    \label{eq:sectors}
    && \mathcal{S}_1 = \left\{(x,y,z)\in\mathbb{R}^3 : z\le h-r,\ x\le -d/2\right\}, \nonumber \\
    && \mathcal{S}_2 = \left\{(x,y,z)\in\mathbb{R}^3 : z>h-r, \ x<-d+r \right\}, \nonumber \\
    && \mathcal{S}_3 = \left\{(x,y,z)\in\mathbb{R}^3 : z>h-r, \ -d+r \le x \le -r \right\}, \nonumber \\
    && \mathcal{S}_4 = \left\{(x,y,z)\in\mathbb{R}^3 : z>h-r, \ x> -r \right\}, \nonumber \\
    && \mathcal{S}_5 = \left\{(x,y,z)\in\mathbb{R}^3 : z\le h-r, \ x > -d/2 \right\}.
\end{eqnarray}
% }
where $h$ is the drone fly height (with respect to $\mathcal{F}_I$), $r$ is the radius of the transition arcs, and $d$ is the horizontal separation between $\mathbf{p}_s$ and $\mathbf{p}_f$. Starting and final points, sectors, and variables defined here are illustrated in Figure~\ref{fig:sectors}.

\subsection{Localization}
\vspace{-0.0cm}

The Onboard-SDK-ROS package provides a georeferenced estimate of the drone's global orientation and position (DJI-SDK Pose). This information comes from the sensory fusion of the available GPS, 9DoF IMU, and barometer, which results in an accuracy of approximately $2\mathrm{m}$. Such an estimate is good enough when on a cruise flight and was therefore used throughout the flight. However, this may be insufficient when landing on a 1x1 \textrm{m} platform like the one proposed in this work, whose dimensions are less than the position estimate's accuracy. Besides, if the landing pad position changes or the georeference is not precise enough, the drone might not be able to land at the right location using GPS localization alone.

For these reasons, it is necessary to obtain additional information that allows the improvement of the drone's estimated position with respect to the landing site.
This paper presents a sensor fusion strategy to improve localization by merging the DJI-SDK pose estimation with information from: (i) an ArUco marker detection technique; and (ii) multilateration using Ultra-wideband (UWB) communication devices.

% This paper presents two approaches to this: (i) use of an ArUco marker; (ii) use of Ultra-wideband (UWB) communication devices in the landing area. \textcolor{red}{(iii) adicionar aqui a fusão de ArUco e UWB?} 

\subsubsection{ArUco}
\label{sec:ArUco}

% \victor{
For the use of the marker detection technique for localization, ArUco markers were printed and placed on top of the landing platform.
% }
% For the use of ArUco, a marker of size 0.8x0.8\textrm{m} was printed and placed on top of the landing platform.
%
A camera attached to the drone, pointing downwards, provides images of the marker during landing. ArUco markers have features that facilitate their identification in the image, such as well-defined borders and high color contrast. In addition, the markers do not present ambiguities in their orientation.

%are created in a way that there are not 
%, Which facilitates the estimation of the camera's pose in relation to the marker.

Thus, specific OpenCV algorithms identify the ArUco, as well as estimate the relative pose of the marker with respect to the camera. This last step is done by solving the problem of \textit{PnP (Perspective-n-Point)}, which proposes to estimate the three-dimensional pose of a calibrated camera given a set of 3D points and their corresponding 2D projections on the camera plane.

% Knowing the actual size of the marker and the intrinsic calibration parameters of the camera, it is possible to find the pose that minimizes the projection errors of the points on the camera plane.
It is possible to find the pose that minimizes the projection errors of the points on the camera plane by knowing the actual size of the marker and the intrinsic calibration parameters of the camera. These points must be distinguishable from each other, and in this case, the corners of the ArUco and its orientation allow for differentiating each of its four corners before sending to a PnP solver algorithm~\citep{pnp_algorithm2,pnp_algorithm3}.
% These points must be distinguishable from each other, and in this case, the corners of ArUco are used to solve the PnP problem. As stated earlier, it is possible to determine the orientation of the marker, which allows differentiating each of its four corners. Thus, algorithms that solve the problem of PnP~\citep{pnp_algorithm2,pnp_algorithm3} can be used.

% \textcolor{red}{Incluir trabalhos que utilizam uma combinação diferente de marcadores.}

\begin{figure}[t]
      \vspace{-0.0cm}
      \centering
      \includegraphics[width=0.4\linewidth]{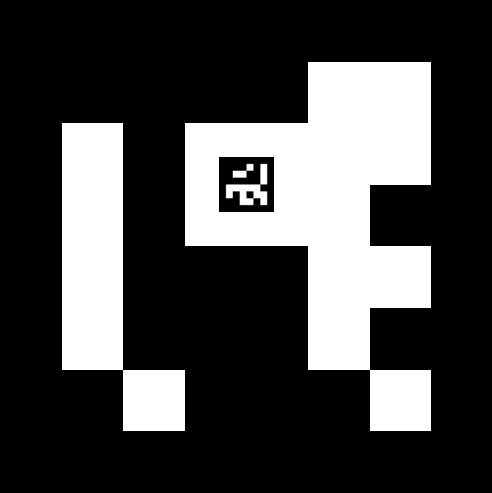}
      \vspace{-0.0cm}
      \caption{Modified ArUco marker. An smaller marker is placed in a bigger one.}
      \label{fig:double_aruco}
      \vspace{-0.5cm}
\end{figure}

There are works in the literature that address strategies to improve the detection of the markers. A common method is to merge different AR markers to create a group that provides a better pose estimation, minimizing noise and occlusion, as presented in \citep{de2019vision}.
Large ArUco markers can be detected from high altitudes. However, when the drone is approaching the platform, this ArUco is quickly lost by the camera.
% \adriano{
% A large ArUco marker has the advantage of being detectable from high altitudes. However, when the drone is approaching the platform, the larger ArUco quickly stays out of the camera.
On the other hand, a smaller ArUco has the advantage of being detectable when the drone is close to the platform (if there is no high horizontal error), even though it is difficult to be detected at high altitudes. %. Although, at high altitudes, it is difficult to be detected.
%
% Intending to detect the platform at high and low altitudes, we considered a modified ArUco marker that has a smaller marker (0.09x0.09\textrm{m}) inside a larger one (0.8x0.8\textrm{m}). %
To improve the marker detection range at high and low altitudes, we considered a modified ArUco marker that has a smaller marker (0.09x0.09\textrm{m}) inside a larger one (0.8x0.8\textrm{m}). %
Figure~\ref{fig:double_aruco} shows this modified ArUco.
%
% \victor{
The inclusion of the smaller marker may harm the detection of the larger one. Nonetheless, in our tests,  this problem did not occur. 
% }
%

\color{black}   

\subsubsection{Ultra-Wideband (UWB) Devices}

Although the ArUco marker detection provides a good pose estimation, the method is not robust for detection in low-light environments or under visual occlusion situations.
% \victor{
For this reason, we consider using another localization based on ultra-wideband devices, which works on these conditions and increases the landing strategy robustness.
% }

Devices based on ultra-wideband wireless technology are commonly used for low-energy IoT communication or localization. This technology uses radio waves with a bandwidth greater than $500 \mathrm{MHz}$, which reduces the loss due to obstructions and reflections of the environment, consequently increasing the security of transmissions~\citep{sahinoglu2008ultra}.

UWB-based localization systems can be used indoors and outdoors, with an accuracy of up to $20 cm$, according to some manufacturers. In this method, multilateration algorithms estimate the position $x_T$, $y_T$ and $z_T$ of a mobile device (called tag) with respect to a fixed reference, where other devices (called anchors) are located.

In this paper, we use Decawave DWM1001\footnote{Decawave - \href{https://www.decawave.com/product/dwm1001-development-board/}{https://www.decawave.com/product/dwm1001-development-board/}} UWB devices to estimate the position of the drone with greater precision when approaching the landing platform. A minimum of five devices are required for the algorithm to work, one tag embedded in the drone and four anchors in known positions; one of these devices is set as the base anchor.

The position is calculated based on the distance of the tag with respect to the anchors, which comes from the Time Difference of Arrival (TDoA) of the transmitted signal, multiplied by the speed of signal propagation (speed of light), as presented in~\citep{sayed2005network}. 
%

% \victor{
Consider a set of enumerated UWB devices, where index $0$ represents the tag placed on the robot, index $1$ the base anchor (or main anchor), and the higher indexes the other anchors used on the system. The distance $d_{i1}$ from the $i$-th anchor to the base anchor is given by:
\begin{align}
    \vspace{0.4cm}
    &d_{i1} = (t_i - t_1)c,\ i=2,...,N,
    \label{eq:TDoa}
    \vspace{0.4cm}
\end{align}
% where $d_{i1}$ is the distance between anchor $i$ and the base anchor,
where $t_i$ is the instant of time the signal sent by the tag reaches anchor $i$, and $t_1$ is the instant of time this signal reaches the base anchor. Light speed is $c$ and the number of anchors is $N$, such that $N\geq4$.
% }
Distances in equation~\eqref{eq:TDoa} result on an intersection region that represents the Tag's position, obtained as the solution of the following set of equations:
\begin{equation}
    \begin{aligned}
        \label{eq:TDoA_eq}
        &(d_{21}^2 + d_1^2)^2 = (x_2 + y_2 + z_2)^2 - 2J_2 + d_1^2,\\
        &(d_{31}^2 + d_1^2)^2 = (x_3 + y_3 + z_3)^2 - 2J_3 + d_1^2,\\
        &\vdots \\
        &(d_{N1}^2 + d_1^2)^2 = (x_{N} + y_{N} + z_{N})^2 - 2J_N + d_1^2,
    \end{aligned}
    \vspace{0.1cm}
\end{equation}
%
%
% \victor{
where $J_i = (x_ix_{T} + y_iy_{T} + z_iz_{T})$ and $d_1$ is the distance from the tag to the base anchor. Considering $t_0$ the instant of time the signal is sent by the tag, $d_1$ can be computed as follow:
% }
%
% where $J_i = (x_ix_{T} + y_iy_{T} + z_iz_{T})$ and $d_1$ is obtained through the equation below, considering $t^0$ the instant of time the signal is sent by the tag and $t_1$ the instant that this signal reaches the base anchor:
%
\begin{equation}
    d_{1} = (t_1 - t_0)c.
\end{equation}

To use Decawave devices, a maximum distance of $10 \textrm{m}$ must be kept between the anchors and the tag. This methodology does not estimate orientation, whereas the ArUco estimation does.
%A Figura \ref{fig:TDoA_diagram} ilustra um diagrama esquemático do método.

\color{black}

\subsubsection{Sensory Fusion}
\label{sec:fusao_sensorial}

One way to improve the localization and more accurately estimate the drone's position, orientation, and speed states, is to use information from several sensors. Sensory fusion methods use data from different devices to obtain more accurate information on the states of interest. For instance, the software on the DJI Matrice 100 uses data from the GPS, 9DoF IMU, and barometer to provide the drone's pose and speeds (DJI-SDK pose).

One of the most common fusion methods is the Extended Kalman filter (EKF)~\citep{thrun2000probabilistic}. In this sense, the PnP estimation of the ArUco's marker location and the UWB localization are merged with the DJI-SDK pose data to allow a more accurate landing. As the sensors provide data referring to coincident states, a bias is considered in the position estimated by GPS.
% \textcolor{red}{As the sensor data transmit information referring to coincident states, a bias is considered in the estimated position by the ArUco detection or by the UWB localization system.}

It is possible to divide Kalman's Extended fusion and filtering method into two stages: Prediction and Correction. For simplicity, the equations are presented with the notation $b\gets a$ indicating that $b$ is updated with the value of $a$.

The prediction step in the discrete EKF involves the state vector $\bar{\mathbf{x}}$ and the covariance matrix $P$:
\begin{eqnarray}
    \bar{\mathbf{x}} &\gets& f(\bar{\mathbf{x}},\mathbf{u},\Delta t), \label{eq:prop_x} \\
    P &\gets& FPF^T + GQ_uG^T + Q_f,
    \label{eq:prop_P}
\end{eqnarray}
%
%\textcolor{red}{O 'P' do lado direito da equacao (matriz de covariancia) nao deveria ser representado como P', algo assim? para diferenciar do lado esquerdo e dizer que esta sendo atualizada?}
where $f$ represents the state propagation model, which involves current estimate $\bar{\mathbf{x}}$, the input vector $\mathbf{u}$ and the timestep $\Delta t$. Matrix $F \equiv F(\bar{\mathbf{x}},\mathbf{u},\Delta t)$ is the partial derivative of $f$ with respect to $\bar{\mathbf{x}}$, and matrix $G$ is the partial derivative of $f$ with respect to $\mathbf{u}$. Matrix $Q_u$ is the covariance matrix associated with input vector $\mathbf{u}$, and $Q_f$ is a covariance matrix associated with the model.

The correction step is defined as follows:
\begin{eqnarray}
    \bar{\mathbf{x}} &\gets& \bar{\mathbf{x}} + K\left(\mathbf{w}-h(\bar{\mathbf{x}})\right), \label{eq:update_x} \\
    P &\gets& \left(I-KH\right)P, \label{eq:update_P}
\end{eqnarray}
where $\mathbf{w}$ is the measurement vector and $h(\bar{\mathbf{x}})$ is the measurement model, which represents the expected value of $\mathbf{w}$ given the current estimated state $\bar{\mathbf{x}}$. Matrix $H$ is the Jacobian of  $h(\bar{\mathbf{x}})$, and $I$ is an identity matrix. Finally $K$ represents the Kalman gain, given by:
\begin{equation}
    \label{eq:Kalman_gain}
    K = PH^T\left(HPH^T+R\right)^{-1},
\end{equation}
where $R$ is the covariance of the measurement data $\mathbf{w}$.

The strategy defines the input vector as $\mathbf{u} = [\mathbf{u}_v^T\ \mathbf{u}_\omega^T]^T = [v_x \ v_y \ v_z \ \omega_x \ \omega_y \ \omega_z]^T$, where $v_x$, $v_y$ and $v_z$ are the linear velocities of the drone with respect to the world frame, and $\omega_x$, $\omega_y$, and $\omega_z$ are the angular velocities in the body frame. In the correction steps, the measurement $\mathbf{w}$ may assume three different values: (i) position and Euler angles from the DJI-SDK pose; (ii) position and Euler angles from the ArUco-PnP; and (iii) position from the UWB system.

In general, the data collected by DJI-SDK position (GPS), compared to the ArUco and UWB, have an unknown non-zero average error. Therefore, the direct use of these measurements in the correction step causes an oscillation in the estimation. Despite containing a displacement, the data from DJI-SDK is the only available throughout the whole experiment and cannot be discarded.
In order to merge the data from these two sensors properly, we consider extra states representing a bias associated with the DJI-SDK position (GPS) measurements with respect to the landing platform's location (ArUco or Decawave). In fact, the proposed filter incorporates the following $12$ states:
%
% \begin{equation}
%     \label{eq:vector_of_states}
%     \bar{\mathbf{x}} = \left[\bar{\mathbf{p}}\ \bar{\mathbf{p}}_b\right]^T= [\underbrace{\bar{x} \ \bar{y} \ \bar{z}}_{\mbox{\small{position}}} \ \underbrace{\bar{b}_x \ \bar{b}_y \ \bar{b}_z}_{\mbox{\small{\textit{bias}}}}]^T,
% \end{equation}
\begin{equation}
    \label{eq:vector_of_states}
    \bar{\mathbf{x}} = %\left[\bar{\mathbf{p}}\ \bar{\mathbf{r}}\ \bar{\mathbf{b}}_p\ \bar{\mathbf{b}}_\omega\right]^T =
    [\underbrace{\bar{x} \ \bar{y} \ \bar{z}}_{\mbox{\small{position}}} \ \underbrace{\bar{\phi} \ \bar{\theta} \ \bar{\psi}}_{\mbox{\small{angles}}} \ \underbrace{\bar{b}_x \ \bar{b}_y \ \bar{b}_z}_{\mbox{\small{bias GPS}}} \ \underbrace{\bar{b}_{\omega x} \ \bar{b}_{\omega y} \ \bar{b}_{\omega z}}_{\mbox{\small{\textit{bias gyro}}}}]^T,
\end{equation}
\noindent where $\bar{\mathbf{p}} \equiv [\bar{x} \ \bar{y} \ \bar{z}]^T$ is the drone's position, $\bar{\mathbf{r}} \equiv [\bar{\phi} \ \bar{\theta} \ \bar{\psi}]^T$ the drone's orientation in %with
Euler angles, $\bar{\mathbf{b}}_p \equiv [\bar{b}_x \ \bar{b}_y \ \bar{b}_z]^T$ is the GPS bias, and $\bar{\mathbf{b}}_\omega \equiv [\bar{b}_{\omega x} \ \bar{b}_{\omega y} \ \bar{b}_{\omega z}]^T$ is the drone's gyro bias. More precisely, $\bar{\mathbf{b}}_p$ is the GPS bias with respect to the (pre-defined) location of the landing platform. Note that the filter does not have bias states related to the ArUco orientation. For this reason, the marker geographical orientation must be known with relative precision.

The propagation model $f(\bar{\mathbf{x}},\mathbf{u},\Delta t)$ is given by:
% \begin{equation}
% \label{predicao}
% f(\bar{\mathbf{x}},\mathbf{u},\Delta t) = \left[\begin{array}{cc}
% &\bar{\mathbf{p}} + \mathbf{u}\Delta t\\
% &\bar{\mathbf{p}}_b
% \end{array}\right].
% \end{equation}
\begin{equation}
\label{predicao}
% f(\bar{\mathbf{x}},\mathbf{u},\Delta t) = \left[\begin{array}{cc}
% &\bar{\mathbf{p}} + \mathbf{u}_v\Delta t\\
% &\bar{\mathbf{r}} + J_{r}(\mathbf{u}_\omega{-}\bar{\mathbf{b}}_\omega)\Delta t\\
% &\bar{\mathbf{b}}_p\\
% &\bar{\mathbf{b}}_\omega
% \end{array}\right].
f(\bar{\mathbf{x}},\mathbf{u},\Delta t) = \left[\begin{array}{c}
\bar{\mathbf{p}}\\
\bar{\mathbf{r}}\\
\bar{\mathbf{b}}_p\\
\bar{\mathbf{b}}_\omega
\end{array}\right] + \left[\begin{array}{c}
\mathbf{u}_v\\
J_{r}(\mathbf{u}_\omega{-}\bar{\mathbf{b}}_\omega)\\
0\\
0
\end{array}\right]\Delta t,
\end{equation}
in which $J_r \equiv J_r(\bar{\phi},\bar{\theta})$ is the Jacobian matrix that transforms the angular velocities $\mathbf{u}_\omega$ in the derivatives of the Euler angles $\bar{\mathbf{r}}$. Note that the prediction model assumes constant bias states.

The filter considers three measurement models. The first one represents the expected measurement of position and orientation data provided by the DJI SDK. In this correction step, there is a binary variable that determines whether the landing platform local data is already being collected or not. This variable $\xi$ determines the inclusion of bias in the GPS measurement model. This variable is $1$ when the local data has already been observed and $0$ if not. Thus, we have:
\begin{equation}
    \label{eq:measurement_model_sdk}
    h_{SDK}(\bar{\mathbf{x}}) = \left[\begin{array}{c}
        \bar{\mathbf{p}} + \xi\bar{\mathbf{b}}_p  \\
        \bar{\mathbf{r}} 
    \end{array}\right].
\end{equation}

Let $H_c^d\in SE(3)$ be a constant homogeneous matrix that represents the pose of the camera with respect to the drone and $H_{a}^w\in SE(3)$ be a constant homogeneous matrix that represents the pose of the ArUco in the world. The matrix $\bar{H}_d^w \equiv \bar{H}_d^w(\bar{\mathbf{p}},\bar{\mathbf{r}})$ represents the pose of the drone in the world frame. Then, the expected pose of the ArUco with respect to the camera (data provided by the PnP algorithm) can be written as:
\begin{equation}
    \bar{H}_a^c = (H_c^d)^{-1} (\bar{H}_d^w)^{-1} H_{a}^w.
\end{equation}

 The matrices $H_c^d$ and $H_{a}^w$ are known \textit{a priori}, whereas the matrix $\bar{H}_d^w$ is obtained from the filter states $\bar{\mathbf{p}}$ and $\bar{\mathbf{r}}$.
Since the filter works with Euler angles, the measurement model of the ArUco information is given by:
\begin{equation}
    \label{eq:measurement_model_aruco}
    h_{ArUco}(\bar{\mathbf{x}}) = \left[\begin{array}{c}
        \mbox{\textit{get\_pos}}(\bar{H}_a^c)  \\
        \mbox{\textit{get\_euler}}(\bar{H}_a^c)
    \end{array}\right],
\end{equation}
\noindent in which \textit{get\_pos()} and \textit{get\_euler()} are functions that return the position and the Euler angles associated to a given homogeneous transformation matrix.

% \adriano{
It is important to comment about the reason we feed the filter with the original pose of the ArUco (direct from PnP, with no transformation) with respect to the camera. A different strategy would consider a sequence of homogeneous transformations in order to obtain the pose of the drone with respect to the world and feed the filter with this transformed data. In this case, model $h_{ArUco}(\bar{\mathbf{x}})$ would correspond to the identity function. The justification for the choice of our strategy relies on the noise levels of the ArUco measurement. In fact, the position and the yaw angle of the marker with respect to the camera are estimated with good precision, while the roll and pitch have significantly higher uncertainty. In our approach, we are able to provide these levels of accuracy in a diagonal covariance matrix $R$. Using a sequence of homogeneous transformations, the high noise in the roll and pitch reflects on the computed position of the drone, and consequently, would worse the filter response.
%
% When the sequence of homogeneous transformations is used, the high noise in the roll and pitch is reflected also on the computed position of the drone, and consequently, would worse the filter response. 
The consideration of the original pose measurement allows the filter to properly treat the signals according to their correct covariances.
% }

Finally, since the UWB system only provides a position measurement, its measurement model is given by:
\begin{equation}
    \label{eq:measurement_model_uwb}
    h_{UWB}(\bar{\mathbf{x}}) = \bar{\mathbf{p}}.
\end{equation}

In this way, the estimation of the drone's pose improves, increasing safety and accuracy during landing. The results in Section~\ref{sec:results} demonstrate the effectiveness of the methodology.

%##########################
%##########################
%#########################

\subsection{Control with Vector Fields}

The quadrotor control is divided into two levels: high and low. The main objective is to enforce the drone to follow the path presented in Section \ref{sec:path_planning}. In order to accomplish this task, the proposed high-level controller is an artificial vector field created to allow path-following assuming the drone behaves as a simple integrator. The low-level controller can be any controller able to impose this desired vector field-based velocity behavior to the real system. For instance, in the experiments presented in this work, the drone uses the low-level controller proposed in \cite{9196605}. Next, we describe how the desired vector field can be defined by means of the strategy proposed in \citep{goncalves2010}.

Traditional control techniques, or even flight modes that use waypoints on the map could have been used, however with disadvantages.
In the proposed technique, the generated path is smooth, eliminating unwanted effects caused by switching between control laws at each waypoint. In addition, paths such as the ones in sectors $S_2$ and $S_4$  (Figure~\ref{fig:sectors}) can be optimized to generate less abrupt movements with the load.%, by adjusting parameters.

The control by vector fields is used to follow paths, not trajectories. % As will become clear below,
The proposed control law is a function only of the drone's state $\mathbf{p}$, therefore, it does not directly depend on time $t$. 
This property is particularly interesting because it does not present two problems associated to trajectory control: (i) if the reference starts very far from the drone's initial position, it may pass through an aggressive transient, which is unwanted; (ii) in the event of a temporary system failure that causes the drone to stop responding for a while, when the drone returns, the reference may be too far away, generating an additional transient state.
% \adriano{acho que aqui esta bom}

%O controle de alto nível lida apenas com velocidades lineares e considera que o modelo na equação \ref{eq:integrator_model} é válido. Assim, a estratégia utilizada envolve a definição de um campo cujas linhas integrais convergem para a curva $\mathcal{C}$ desejada, como visto em \cite{goncalves2010}.

%A curva $\mathcal{C}$ é representada pela interseção de duas superfícies de nível zero, $\alpha_i(\mathbf{p})=0$, com $\alpha_i:\mathbb{R}^3\to\mathbb{R}$, $i=1,2$. Ou seja, $\mathcal{C}=\left\{\mathbf{p}\in\mathbb{R}^3:\alpha_1(\mathbf{p}){=}0 \wedge \alpha_2(\mathbf{p}){=}0\right\}$. A Figura \ref{fig:alphas} ilustra esta representação.

\begin{figure}[t]
      \vspace{-0.0cm}
      \centering
      \includegraphics[width=0.35\textwidth]{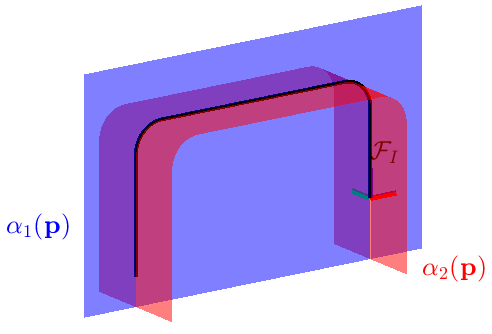}
      \vspace{-0.0cm}
      \caption{Graphical representation of $\alpha_1$ and $\alpha_2$ functions, equations~\eqref{eq:alpha_1} and \eqref{eq:alpha_2}, respectively.}
      \label{fig:alphas}
      \vspace{-0.0cm}
\end{figure}

To represent a curve $\mathcal{C}$, it is necessary to define scalar functions $\alpha_i:\mathbb{R}^3\to\mathbb{R}$, $i=1,2$, for each sector $\mathcal{S}_j$, $j=1,2,3,4,5$, in a way that the intersection of their zero-level surfaces, $\alpha_i(\mathbf{p})=0$, generates the desired curve~\citep{goncalves2010}. This means that $\mathcal{C}$ is defined by  $\mathcal{C}=\{\mathbf{p}\in\mathbb{R}^3:\alpha_1(\mathbf{p}){=}0 \wedge \alpha_2(\mathbf{p}){=}0\}$.
Figure~\ref{fig:alphas} illustrates this representation in the case of the path defined in Section \ref{sec:path_planning}. %, with functions $\alpha_1$ and $\alpha_2$ defined as:
The functions $\alpha_1\equiv\alpha_1(y)$ and $\alpha_2\equiv\alpha_1(x,z)$ are defined as:
\vspace{0.2cm}
\begin{eqnarray}
    \label{eq:alpha_1}
    \alpha_1 = y, %\left\{\begin{array}{l@{}c}
    %     y,  \ \ \ &\mbox{starting from $\mathbf{p}_s$}  \\
    %     -y,  \ \ \ &\mbox{starting from $\mathbf{p}_f$} \\ 
    % \end{array}\right.,
\end{eqnarray}
% \adriano{in which $s=1$ if the drone \luciano{is} going from $\mathbf{p}_s$ to $\mathbf{p}_f$ and $s=-1$ if it is returning after the delivery is made, \emph{i.e.} going from $\mathbf{p}_f$ to $\mathbf{p}_s$. Function $\alpha_2$ is defined as:}
%
%\vspace{-0.5cm}
% {\color{adriano}
\begin{eqnarray}
    \label{eq:alpha_2}
    \hspace{-0.4cm}\alpha_2 {=} \left\{\begin{array}{l@{}c}
        -x-d, \ \ \ &\mbox{if} \ \mathbf{p}\in\mathcal{S}_1 \\
        \sqrt{(x{+}d{-}r)^2+(z{-}h{+}r)^2}-r, \ \ \ &\mbox{if} \ \mathbf{p}\in\mathcal{S}_2 \\
        z-h, \ \ \ &\mbox{if} \ \mathbf{p}\in\mathcal{S}_3 \\
        \sqrt{(x{+}r)^2+(z{-}h{+}r)^2}-r, \ \ \ &\mbox{if} \ \mathbf{p}\in\mathcal{S}_4 \\
        x, \ \ \ &\mbox{if} \ \mathbf{p}\in\mathcal{S}_5
    \end{array}\right.
\end{eqnarray}
% }
\vspace{0.2cm}

This way, for each point in space, a convergent and a tangential component to the curve are given by:
\begin{equation}
    \vspace{0.2cm}
    \centering
    F_{conv} = \frac{\nabla V}{\|\nabla V\|}, \quad\quad F_{tang} = s\frac{\nabla \alpha_1 {\times} \nabla \alpha_2}{\|\nabla \alpha_1 {\times} \nabla \alpha_2\|},
    \label{eq:Fconv_Ftang}
    \vspace{0.2cm}
\end{equation}
\noindent where $V=\frac{1}{2}\alpha_1^2+\frac{1}{2}\alpha_2^2$ is a Lyapunov function and $\times$ denotes the cross product. We consider $s=1$ when the drone is going to delivery the package (moving from $\mathbf{p}_s$ to $\mathbf{p}_f$) and $s=-1$ when it is returning (from $\mathbf{p}_f$ to $\mathbf{p}_s$). The parameter $s$ is responsible for inverting the sense of motion by changing the direction of the tangent component $F_{tang}$.

As seen in~\citep{goncalves2010}, the functions $G \equiv G(V) = -(2/\pi)\arctan(k_f\sqrt{V})$ and $H \equiv H(V) = \sqrt{1-G^2}$ are defined, where $k_f > 0$ is a convergence weight. This functions are part of the vector field $F(\mathbf{p})$ definition, used in the control strategy, as seen next:
\begin{equation}
\label{eq:campo_vetorial}
    F(\mathbf{p}) = v_r (GF_{conv} + HF_{tang}),
    \vspace{0.1cm}
\end{equation}
where $v_r \equiv v_r(\mathbf{p}) > 0$ is the desired robot velocity, defined by $v_r(\mathbf{p})=v_i$ for $\mathbf{p}\in\mathcal{S}_i,\ i=1,2,3,4,5$, i.e.\, the reference velocity for the drone is dependent on the sector it is.

The drone's orientation can be controlled in a way to keep its $\psi$ angle (around $\mathbf{z}$ axis) in $0^{\circ}$ with respect to reference frame $\mathcal{F}_I$. Thus, a reference $\psi_r=0$ is passed to the lower level controller.
%
% \textcolor{red}{In this manner, the robot keeps itself aligned with vector $[\mathbf{p}_f-\mathbf{p}_s]$, which goes from the starting to the final point.}
%
% Thus, the angular velocity $\dot{\psi}$ can be defined as:
%
% \begin{equation}
%     \label{eq:control_yaw}
%     \dot{\psi} = -k_\psi\psi.
% \end{equation}

% \adriano{
% In the experiments in Section \ref{sec:results}, 
% }
%

The delivery task in autonomous mode can be accomplished by following the logic demonstrated in Algorithm~\ref{code:controle}.
%Once the experiment initialized, the drone follows the logic demonstrated in Algorithm~\ref{code:controle} in order to complete the task in autonomous mode.

\begin{algorithm}[htbp]
    \SetAlgoLined
    $delivered \gets false$\\
    $landingDone \gets false$\\
    $taskFinished \gets false$\\
    \While {$landingDone == false$}{
        {\color{black}
        $states \gets getEKFstates()$\\
        $sector \gets getSector(position)$\\
        $cmd \gets VectorFieldController(states, sector)$\\
        $sendCommands(cmd)$\\}
        \If{$position == landingPosition$}{
            $landingDone \gets true$\\
        }
    }
    
    \While {$delivered == false$}{
        $releasePackage()$\\
        $delivered \gets true$\\
        $takeOff()$\\
    }
    
    $taskFinished \gets true$\\
    \caption{Quadrotor high-level control}
    \label{code:controle}
\end{algorithm}

\section{RESULTS}
\label{sec:results}

\begin{figure}[htbp]
      \vspace{-0.0cm}
      \centering                % left, lower, right, upper
      \includegraphics[clip,trim={0in 0in 0in 0in}, width=0.90\linewidth]{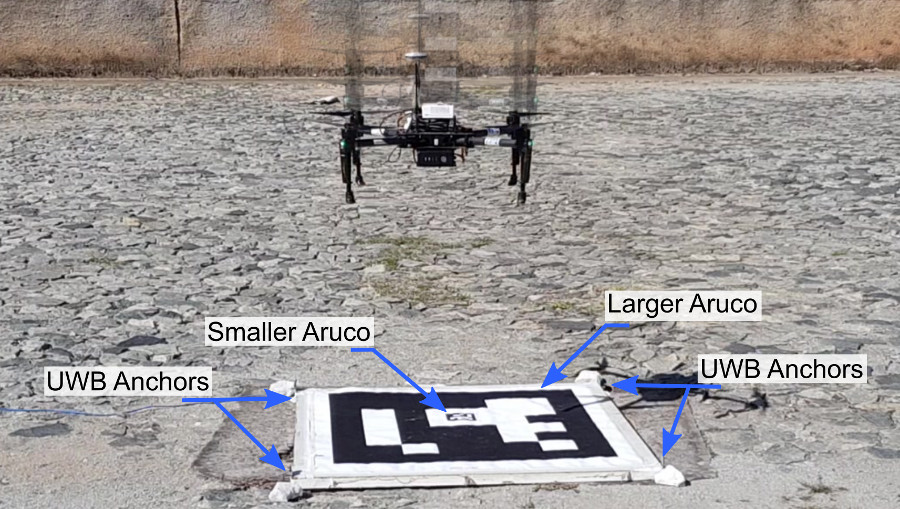}
      \vspace{-0.2cm}
      \caption{Experimental landing setup with planar Aruco markers and UWB tags.}
      \label{fig:expsetupv0}
      \vspace{-0.5cm}
\end{figure}

% \victor{
In this Section, real experimental results are presented, considering a parcel delivery task with a drone in autonomous operation. 
The carried out experiments aim to evaluate the complete autonomous delivery method, as well as the localization strategy in the landing phase. The considered experimental setup is depicted in Figure~\ref{fig:expsetupv0}.

% }
%%%%%%%%%%%%%%%%%%%%%%%%%%%%%%%%%%%%%%%%%%%%%%%%%%%%%%%%%%%%%%%%%%%%%%%%%%%%%%%%%%%%%%%%%%%%%%%%%%%%%%%%%%%%%%%%%%%%%%%%%%%%%%
%

\subsection{Complete Delivery Task Evaluation}

% \subsection{Experimental Validation}

\begin{figure}[ht]
      \vspace{-0.0cm}
      \centering                % left, lower, right, upper
      \includegraphics[clip,trim={1.8in 3.5in 2in 4in}, width=0.5\textwidth]{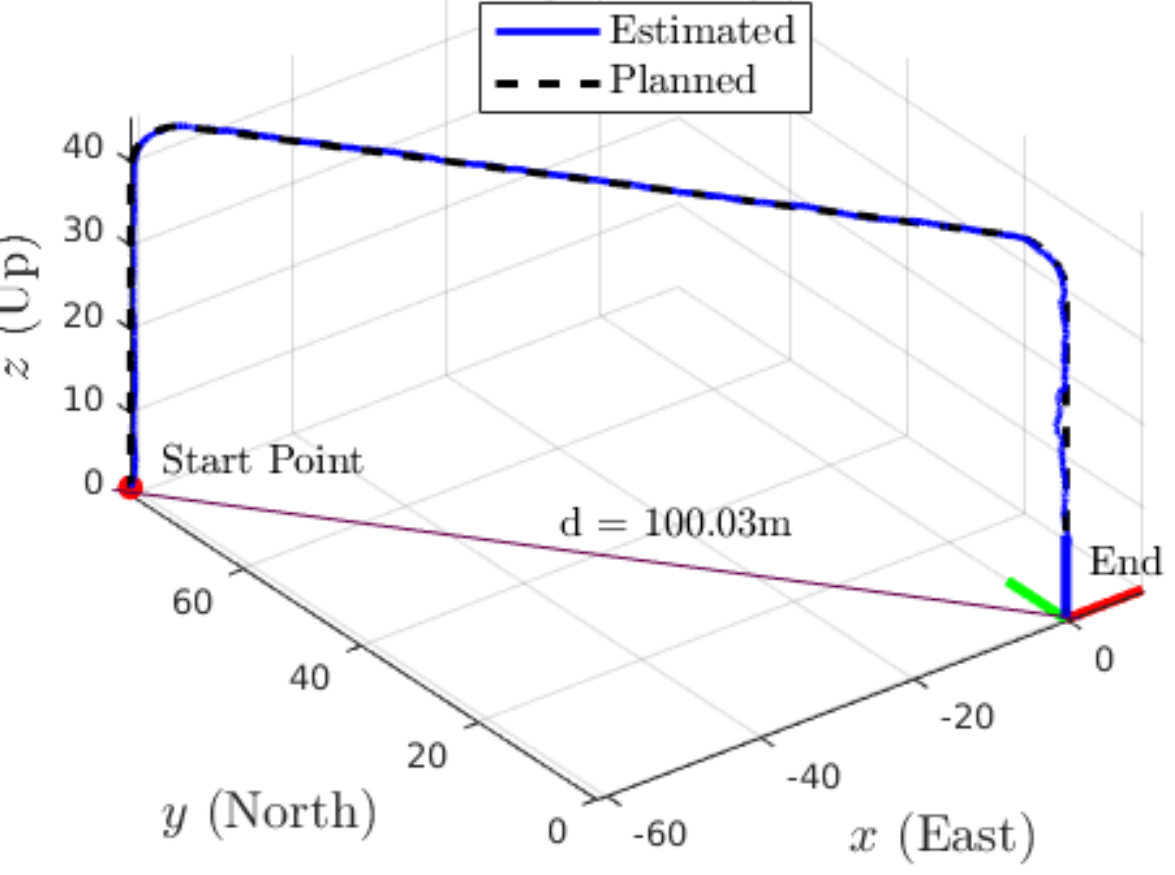}
      \vspace{-0.2cm}
      \caption{Autonomous delivery task results. The performed path is the solid green line and the planned is dashed black.}
      \label{fig:traj_Google_3D}
      \vspace{-0.5cm}
\end{figure}

\begin{figure}[ht]
      \vspace{-0.0cm}
      \centering                % left, lower, right, upper
      \includegraphics[clip,trim={0.2in 3.9in 0.2in 3.9in}, width=0.5\textwidth]{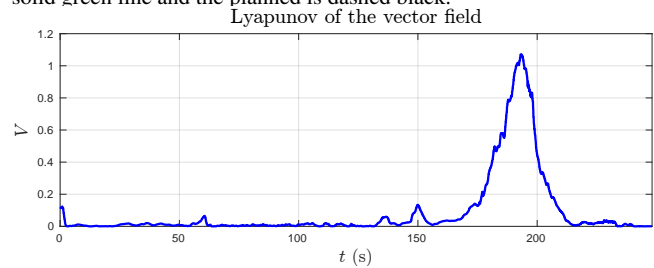}
      \vspace{-0.4cm}
      \caption{Lyapunov function of the vector field strategy. It indicates the distance from the drone to the curve.}
      \label{fig:erroXYZ}
      \vspace{-0.4cm}
\end{figure}

% \victor{
The proposed strategy was validated in a delivery task with the drone following a planned path, from a start point to the endpoint on the landing platform, autonomously.
Figure~\ref{fig:traj_Google_3D} shows a comparison between the path performed by the drone and the planned, using the proposed EKF algorithm with data from the DJI-SDK, and including ArUco markers and UWB on landing. The experiment considers a height $h = 45 \mathrm{m}$ and arcs of the circumference with radius $r = 6 \mathrm{m}$. The parameter $d$ is computed as the distance of drone at its initial position and the expected position of the platform, and in the experiment, $d = 100.03 \mathrm{m}$. Figure~\ref{fig:erroXYZ} illustrates the Lyapunov function of the vector field control, which indicates the distance between the drone and the planned path according to the EKF estimation. This distance increases after $150 \mathrm{s}$ because of the corrections that start when the platform is detected. These changes show to the system that it is not exactly on the path as expected before. In sequence, this distance decreases again with the actions of the controller. As shown in Figure~\ref{fig:traj_Google_3D}, these corrections happened in sector $\mathcal{S}_5$ at the high of approximately $30 \mathrm{m}$. Although the camera detected the marker from $45 \mathrm{m}$ of height, given the high uncertainties in these altitudes, the filter only considers measurements with height below $30 \mathrm{m}$.

%

%

% \adriano{As can be seen in Figure \ref{fig:traj_Google_3D}, this correction happened in sector $\mathcal{S}_5$ at the high of approximately $30 \mathrm{m}$.} \adriano{In fact, the camera detected the marker from $45 \mathrm{m}$ of height. However, given the high uncertainties in these altitudes, the filter only consider measurements with highs below $30 \mathrm{m}$.}
% }
%

% \textcolor{red}{Due to the restrictions imposed by the Covid-19 pandemic, the experiments with the physical platform focused only on the localization system evaluation for landing, which is the main contribution of this paper. Our early version paper has shown experiments considering the complete delivery tasks using autonomous drones \citep{rocha2020desenvolvimento}.}

\begin{figure}[htpb]
      \vspace{-0.0cm}
      \centering
      \includegraphics[clip,trim={0.0cm 0.0cm 0.0cm 0.0cm}, width=0.5\textwidth]{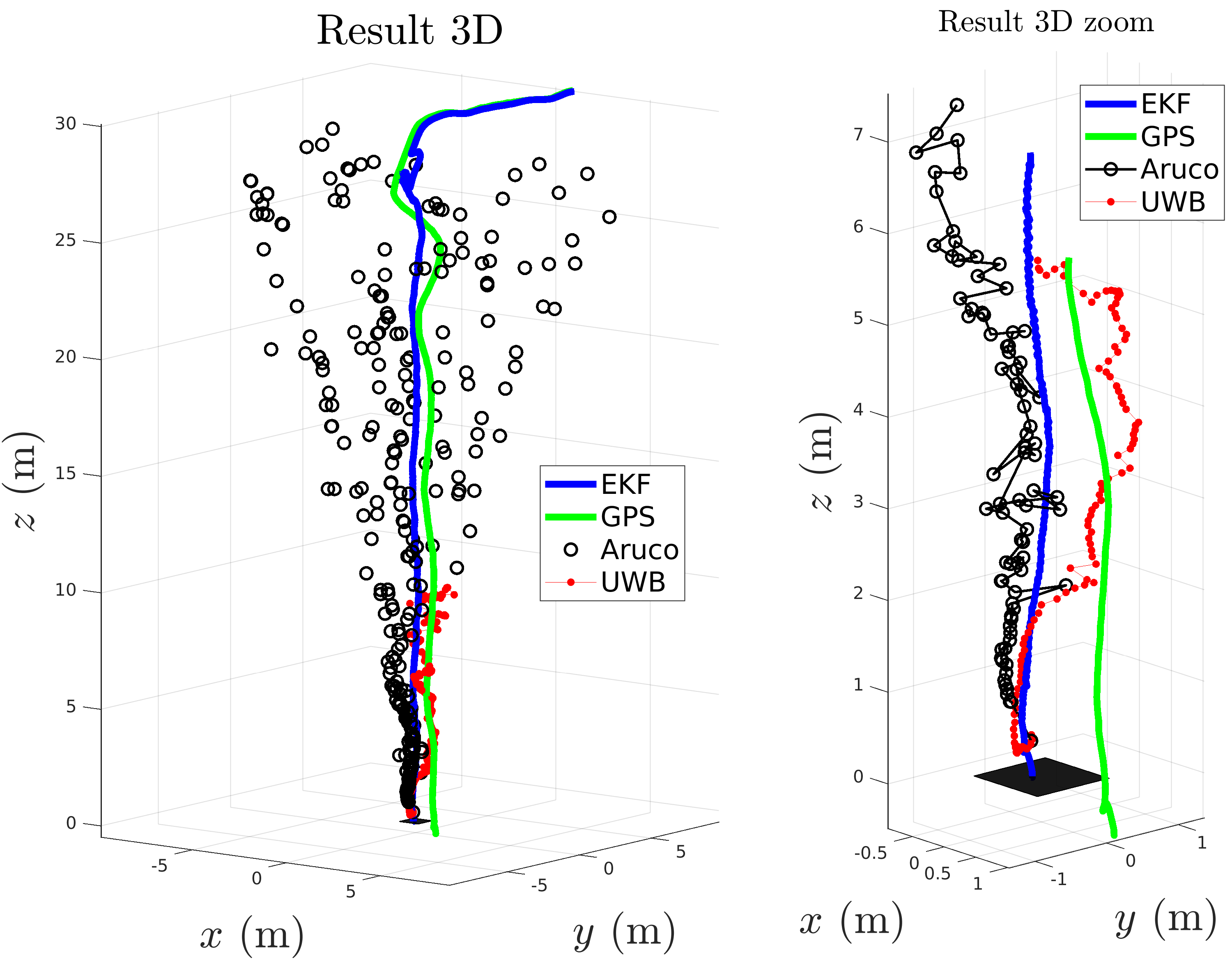}
      \vspace{-0.3cm}
      \caption{Results of 3D position estimation in landing.}
      \label{fig:3d_realExp}
      \vspace{-0.5cm}
\end{figure}

\begin{figure}[htpb]
      \vspace{-0.0cm}
      \centering
      \includegraphics[clip,trim={0.0cm 2.7in 0.0cm 2.7in}, width=0.5\textwidth]{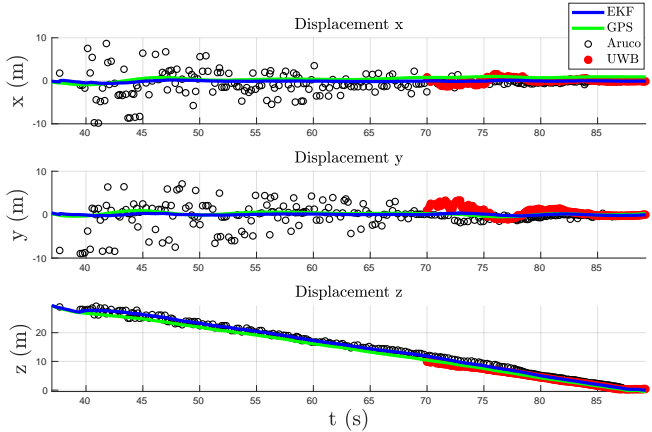}
      \vspace{-0.5cm}
      \caption{Results of position estimation in landing for each axis.}
      \label{fig:axis_realExp}
      \vspace{-0.2cm}
\end{figure}

Figure~\ref{fig:3d_realExp} illustrates the results of the drone's position estimation in another experiment. It shows the data obtained from each sensor separately and from the proposed EKF fusion algorithm during the landing phase. It was a local experiment with $r=2\mathrm{m}$ and $h=30\mathrm{m}$ focused on the landing phase.
In order to plot the position of the ArUco, we considered a direct computation of the position of drone with respect to the frame $\mathcal{F}_I$ (see Figure~\ref{fig:sectors}). It is important to emphasize that this transformed data was not supplied to the filter (see Section \ref{sec:ArUco}). There is considerable noise associated with the ArUco and UWB estimates. These noises are directly proportional to the distance between the drone and the platform, as we can observe in Figure~\ref{fig:axis_realExp}. Besides that, the UWB sensors' information starts to be computed only around $10 \mathrm{m}$ of distance from the landing base.
The system was able to make the drone land almost in the center of the platform, position $[-0.072\ -0.103]^T \mathrm{m}$ in the filter's estimation and $[-0.03\ -0.11]^T\mathrm{m}$ in the ground truth measure. Thus, the distance error from the filter estimation to the ground truth is $0.043\mathrm{m}$, and the ground truth distance to the center of the platform is $0.114\mathrm{m}$.
% }
%

\begin{figure}[htpb]
      \vspace{-0.0cm}
      \centering
    \includegraphics[clip,trim={0.23in 5.5in 0.5in 2.7in}, width=0.5\textwidth]{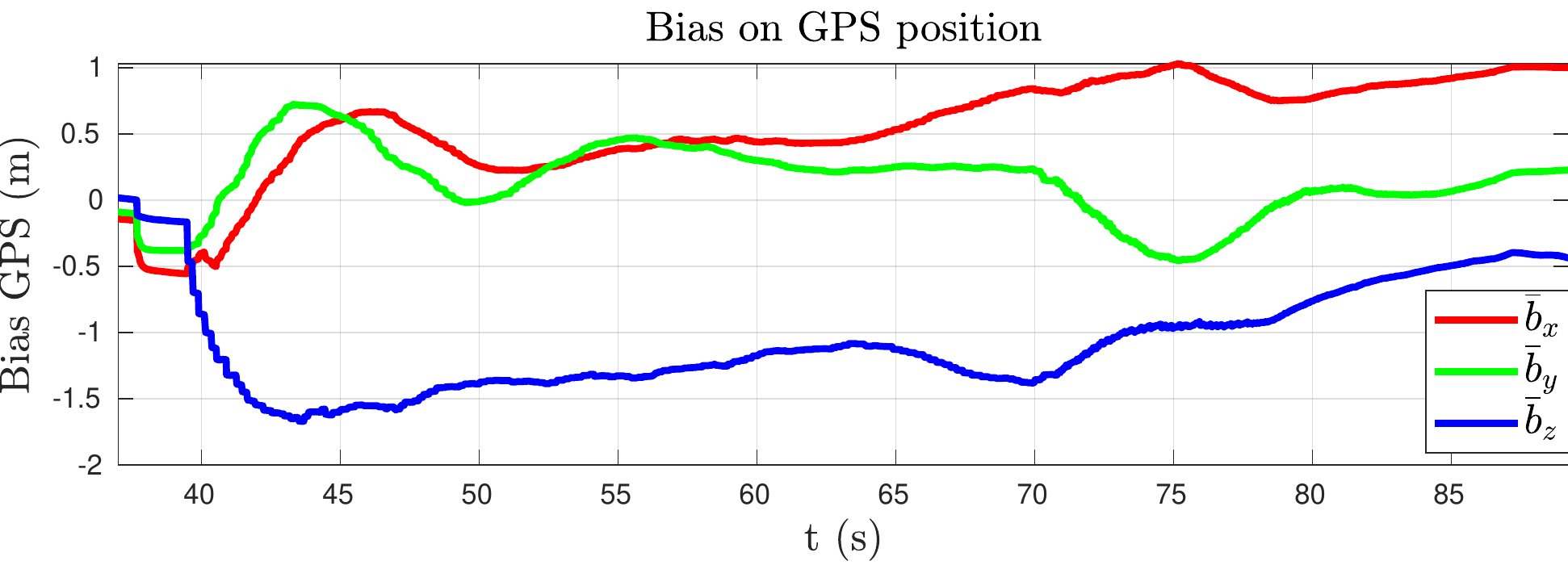}
      \vspace{-0.5cm}
    %   \caption{Bias estimation for GPS position and gyroscope.}
      \caption{Bias estimation for GPS position.}
      \label{fig:bias_realExp}
      \vspace{-0.4cm}
\end{figure}

Note in Figure \ref{fig:3d_realExp} that, despite the noisy data from the ArUco and the UWB, the filter was able to estimate a clear trajectory for the drone. Also, the GPS (from the DJI-SDK) signal has a shift. According to this measurement, the drone did not land correctly on the platform. The filter bias is responsible to correct the GPS shift while neglecting the noise in the ArUco and UWB measurements.
The bias estimated by the filter is depicted in Figure \ref{fig:bias_realExp}.

\begin{figure}[htpb]
      \vspace{-0.0cm}
      \centering
      \includegraphics[clip,trim={0.7cm 7.6cm 0.7cm 7.5cm}, width=0.48\textwidth]{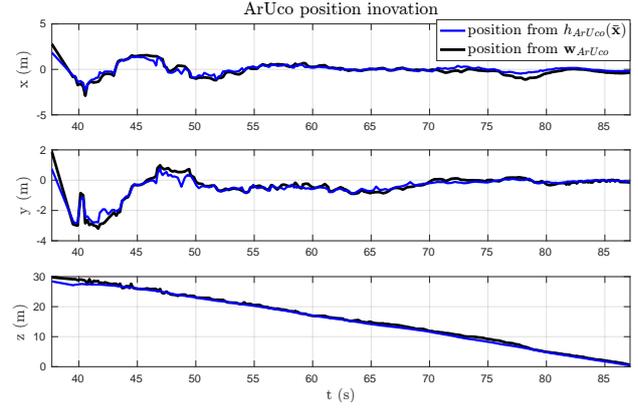}
      \vspace{-0.5cm}
    %   \caption{3D Results of ArUco Detection estimation in landing.\adriano{IMPROVE THIS}}
      \caption{Comparison of the position measurement of the ArUco with the position of the measurement model $h_{ArUco}(\bar{\mathbf{x}})$. The original data has significantly less noise.}
      \label{fig:aruco_realExp}
      \vspace{-0cm}
\end{figure}

As commented in Section \ref{sec:fusao_sensorial}, the noise of the ArUco measurement (ArUco frame with respect to the camera frame) is significantly larger in the estimated roll and pitch angles. When homogeneous transformations are applied to this data to estimate the drone's position, the high noise in the roll and pitch manifest on the $x$ and $y$ position of the drone, as observed in Figures \ref{fig:3d_realExp} and \ref{fig:axis_realExp}. Figure \ref{fig:aruco_realExp} presents a comparison of the original position estimation of the ArUco with respect to the camera ($\mathbf{w}_{ArUco}$) with the expected measurement ($h_{ArUco}(\bar{\mathbf{x}})$). This signal has much less noise than the signal observed in Figure \ref{fig:axis_realExp}.
%
%%%%%%%%%%%%%%%%%%%%%%%%%%%%%%%%%%%%%%%%%%%%%%%%%%%%%%%%%%%%%%%%%%%%%%%%%%%%%%%%%%%%%%%%%%%%%%%%%%%%%%%%%%%%%%%%%%%%%%%%%%%%%%

\subsection{Failures Evaluation}

% \subsection{\textcolor{red}{Robustness evaluation}}

% \color{adriano}

\begin{figure}[ht]
      \vspace{-0.0cm}
      \centering % left, lower, right, upper
      \includegraphics[clip,trim={2.0in 2.2in 2.0in 2.2in}, width=0.40\textwidth]{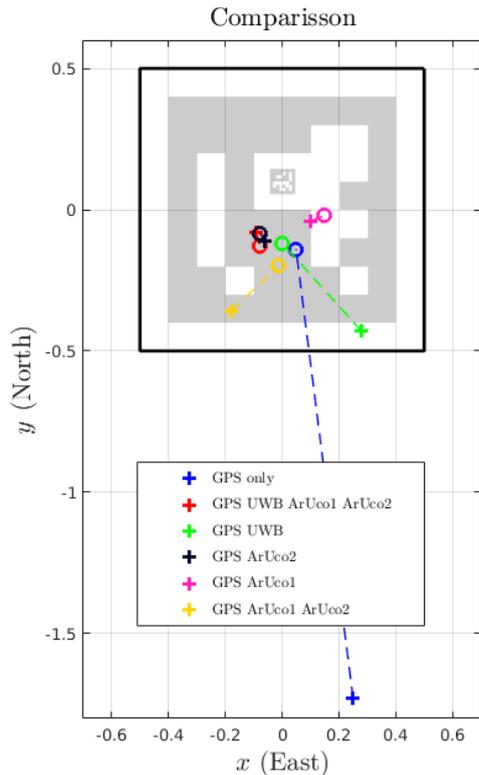}
      \vspace{-0.2cm}
      \caption{Results of $6$ landing experiments that used different combination of the localization strategies.}
      \label{fig:filter_combinations}
      \vspace{-0.0cm}
\end{figure}

The EKF was tested with different combinations of localization methods. The objective is to show that the approach is robust, in the case that if one of the methods fails, the drone still lands on the platform.
Figure \ref{fig:filter_combinations} shows the results of the experiments. The crosses represent the ground truth positions, while the circles represent the estimation of the EKF.
Errors in the ground truth positions are due to the imprecision from both the localization and the controller. As we can see the only method %\textcolor{red}{that could not make the drone land} 
not capable of landing the drone on the platform was the GPS alone. The other strategies consisted of combinations using: UWB, large marker (ArUco1) and small marker (ArUco2).

% \victor{
Still evaluating robustness, we have also conducted another experiment inserting noise on the landing platform position from the defined landing point. Using only the DJI-SDK data as input for the EKF, the drone does not reach the platform, and the platform displacement increases the distance error observed in the previous experiment. 
Considering the inclusion of ArUco detection and the UWB estimation, the drone lands on the platform. In this case, the EKF algorithm considers the displacement on the platform as an error on the GPS data and includes it in the bias estimation.
% }
% Still evaluating robustness, we have also conducted another experiment moving the landing platform $2\textrm{m}$ in the $y$ direction from the defined landing point. The objective is to show that minor errors in the platform position do not affect the landing, considering the proposed localization method. Figure~\ref{fig:platform_displace} shows that even if the endpoint used in the planning has an error concerning the platform's position, the EKF bases on the position of the ArUco and UWB anchors in the landing platform, which causes the drone to land in the correct location. 
% }
% %
% \begin{figure}[ht]
%       \vspace{-0.0cm}
%       \centering % left, lower, right, upper
%       \includegraphics[clip,trim={2.3in 2in 1.9in 1.3in}, width=0.3\textwidth]{images/exp_deslocado_51.pdf}
%       \vspace{-0.0cm}
%       \caption{\textcolor{red}{Results of landing in a platform with $2\textrm{m}$ of displacement in the $y$ direction.}}
%       \label{fig:platform_displace}
%       \vspace{-0.0cm}
% \end{figure}

%%%%%%%%%%%%%%%%%%%%%%%%%%%%%%%%%%%%%%%%%%%%%%%%%%%%%%%%%%%%%%%%%%%%%%%%%%%%%%%%%%%%%%%%%%%%%%%%%%%%%%%%%%%%%%%%%%%%%%%%%%%%%

% \color{adriano}
\vspace{0.2cm}

\subsection{Precision Evaluation}

This section presents a comparison between the landing accuracy of the proposed system. The EKF merging information from all localization systems was considered in $6$ landing experiments. In another $6$ landings, the EKF does not use the ArUco and the UWB data, thus, relying only on the GPS for position estimation.

\begin{figure}[ht]
      \vspace{-0.0cm}
      \centering % left, lower, right, upper
      \includegraphics[clip,trim={1.3in 3.1in 1.3in 3.1in}, width=0.5\textwidth]{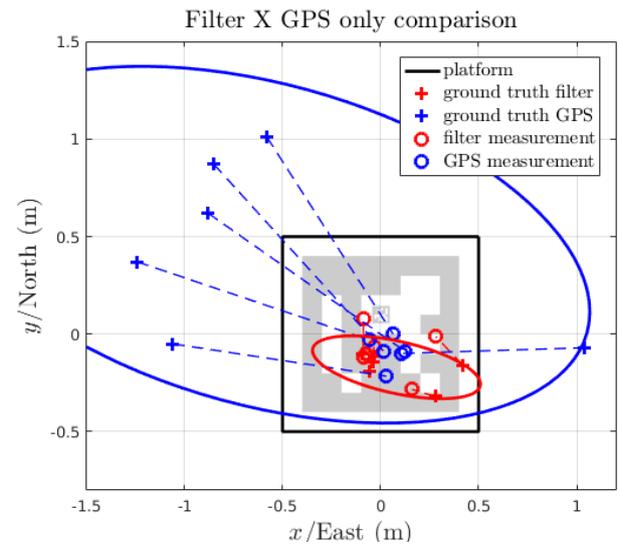}
      \vspace{-0.2cm}
      \caption{Comparison between the system's precision when the ArUco and UWB are considered or not.}
      \label{fig:statistics_12}
      \vspace{-0.0cm}
\end{figure}

Figure \ref{fig:statistics_12} shows the result of these $12$ experiments. The true landing positions correspond to the crosses, while the circles correspond to the position estimation of the filter after the landing. Red represents the data obtained when the filter considers all information, and blue when it uses only the GPS data.
% \adriano{
% The blue ellipse corresponds to the covariance of the landing results using only the GPS and the red to the results with all localization information.
The two ellipses correspond to the covariance of the ground truth landing results and assume $2$ standard deviations.
% }

% \victor{
Note that in none of the experiments trusting only on the GPS, the drone landed on the platform (blue crosses), despite the filter estimated that the drone was close to the platform's center (blue circles). This error was not corrected since that no local information (ArUco or UWB) was available. In these experiments, the distance from the platform's center reached $1.29\mathrm{m}$.
%
% \adriano{
When the filter counted with data from the ArUco and the UWB, the drone landed on the platform all $6$ experiments and the result has a mean distance of $0.19\mathrm{m}$ from the platform's center.
% }
% }
% When the filter counted with the ArUco and the UWB, the drone landed in the platform, with a maximum distance from the platform's center was $0.20\mathrm{m}$.
%

\color{black}

\section{Conclusion and Future Work}
\label{sec:conclusions}

\color{black}
This paper presented a navigation strategy with techniques of path planning, localization, and control for autonomous delivery tasks using drones.
% This paper presented path planning, localization, and vector field control techniques applied in a real delivery tasks using autonomous drones.
Computer Vision algorithms and UWB devices provide pose estimates for the quadrotor to an Extended Kalman Filter, allowing the sensory fusion with the drone's DJI-SDK pose. A vector field-based controller defines commands for the drone to follow a planned path connecting a start point to the landing platform.

Experimental results validate the proposed method in a complete autonomous delivery task. In the drone landing experiments, %
%In this experiment, the drone lands on the platform and,%
it is possible to note the advantage of the proposed localization method over the DJI-SDK pose estimation alone. A robustness analysis shows that the system works in case of failure in one of the localization methods, which can occur in low-light situations, camera vision occlusion, and power supply failure for the UWB devices. 
% \victor{In addition, a test with the platform moved from the known position during planning demonstrated robustness to these situations, with the drone landing on the displaced platform.}
Besides, a precision evaluation compares the accuracy of the landing phase using only the GPS for position estimation and using the proposed EKF with all localization methods, showing the advantages of the adopted strategy. Most of the delivery operations occur in urban and dense populated areas, and all these results show the effectiveness of the adopted method and increase confidence for a safe landing.

Future works include improving safety by detecting obstructions on landing caused by people or animals, for example. Another goal is the increase of the localization precision, for instance, by exploring other sensors and improving the EKF. Options are the consideration of the delay in the pose estimated by the ArUco method and measurement covariance matrices dependent on the vehicle's height.
Besides, a complete $360$-degrees obstacle avoidance system could improve the flight's overall security. We also intend to reach improvements in path planning, considering obstacle avoidance and minimum power consumption. Other future work is to implement fault detection methods using proprioceptive sensors to activate parachutes or other types of security equipment, considering a forced landing. In addition, we intend to investigate situations of landing on a moving platform.

\color{black}

% \adriano{decawave position}

\begin{acknowledgements}

This study was financed in part by the Universidade Federal de Minas Gerais (UFMG), the Vale S.A. and the Instituto Tecnológico Vale (ITV), and the Conselho Nacional de Desenvolvimento Científico e Tecnológico - Brasil (CNPQ) - grant numbers 306286/2020-3 and 315258/2020-9.
An early version of this paper was presented at XXIII Congresso Brasileiro de Automática (CBA 2020).

\end{acknowledgements}

\bibliographystyle{spbasic}      % basic style, author-year citations

\bibliography{refs.bib}   % name your BibTeX data base
% \bibliography{refs.bbl}   % name your BibTeX data base

\end{document}